\documentclass[10pt,twocolumn,letterpaper]{article}

\usepackage{cvpr}              
\definecolor{cvprblue}{rgb}{0.21,0.49,0.74}
\usepackage[pagebackref,breaklinks,colorlinks,allcolors=cvprblue]{hyperref}
\usepackage[misc]{ifsym}
\usepackage{graphicx}
\usepackage{multirow}
\usepackage{subcaption}
\usepackage[table]{xcolor}

\def\paperID{*****} 
\def\confName{CVPR}
\def\confYear{2026}

\title{Mema: Memory-Augmented Adapter for Enhanced Vision-Language Understanding}

\author{
Ying~Liu$^{1}$, Yudong~Han$^{1}$, Kean Shi$^{2}$, Liyuan Pan$^{1,3}$\thanks{Corresponding author} \\
$^{1}$Beijing Institute of Technology,  \\
$^{2}$Peking University, \\
$^{3}$Yangtze Delta Region Academy of Beijing Institude of Technology, Jiaxing, China\\
\texttt{yingliu312@bit.edu.cn},  \texttt{liyuan.pan@bit.edu.cn} \\
}

\begin{document}
\maketitle
\begin{abstract}
Multimodal Large Language Models (MLLMs) have achieved remarkable performance by aligning pretrained visual representations with the linguistic knowledge embedded in Large Language Models (LLMs). However, existing approaches typically rely on final-layer visual features or learnable multi-layer fusion, which often fail to sufficiently exploit hierarchical visual cues without explicit cross-layer interaction design. 
In this work, we propose a Memory-Augmented Adapter (Mema) within the vision encoder. Specifically, Mema maintains a stateful memory that accumulates hierarchical visual representations across layers, with its evolution conditioned on both query embeddings and step-wise visual features. A portion of this memory is selectively injected into token representations via a feedback mechanism, thereby mitigating the attenuation of fine-grained visual cues from shallow layers. Designed as a lightweight and plug-and-play module, Mema integrates seamlessly into pretrained vision encoders without modifying the vanilla backbone architecture. Only a minimal set of additional parameters requires training, enabling adaptive visual feature refinement while reducing training overhead. Extensive experiments across multiple benchmarks demonstrate that Mema consistently improves performance, validating its effectiveness in complex multimodal reasoning tasks. The code have been released at \url{https://github.com/Sisiliu312/Mema}.
\end{abstract}    
\section{Introduction}
\label{sec:intro}

\begin{figure}
    \centering
    \includegraphics[width=\linewidth]{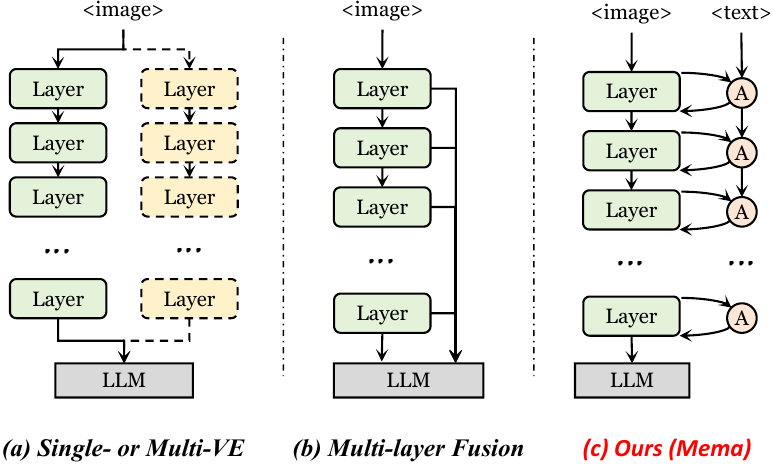}
    \caption{
    Comparison of different strategies for enhancing visual representations in MLLMs. 
    (a) Single- or Multi- vision encoders (VE) rely on representations from fixed layers. 
    (b) Multi-layer fusion methods aggregate features from multiple internal layers. 
    (c) Mema (denoted as A) introduces dynamic, explicit cross-layer interaction via a stateful memory, enabling progressive refinement across layers.
    }
    \label{fig:comp}
\end{figure}

Multimodal Large Language Models (MLLMs)~\cite{llava,Qwen,blip2,dynfocus,ssa,super,lena} have demonstrated remarkable advancements in vision-language reasoning by leveraging powerful pretrained vision encoders~\cite{clip,siglip,vit} and Large Language Models (LLMs)~\cite{Vicuna,Llama,opt}. Vanilla MLLM frameworks primarily utilize the final-layer representations of vision encoders as visual inputs. While computationally efficient, this pipeline predominantly captures high-level semantic information, often overlooking the fine-grained visual cues essential for delicate question answering.
To mitigate this limitation, various strategies have been proposed to enhance fine-grained visual representations. For instance, \citet{pail,internvl,palm-e} adapt scaling laws to vision encoders, enhancing visual perception capabilities by increasing encoder capacity. \citet{mgm,leo,mmvp} integrate multiple heterogeneous vision encoders, leveraging the complementary visual biases embodied in different architectures for collaborative feature encoding, as depicted in Fig.~\ref{fig:comp}(a). While effective, these paradigms often incur substantial computational overhead, hindering their practical scalability. An orthogonal direction focuses on exploiting richer internal representations within a single vision encoder, as depicted in Fig.~\ref{fig:comp}(b). These methods typically aggregate multi-layer representations using techniques such as direct concatenation or adaptive re-weighting~\cite{dc,mmfuser,instruction-guided}.

\begin{figure}[t]
    \centering
    \begin{subfigure}{0.54\linewidth}
        \centering
        \includegraphics[width=\linewidth,height=6cm,keepaspectratio,trim=1 0.25cm 1 0.1cm,clip]{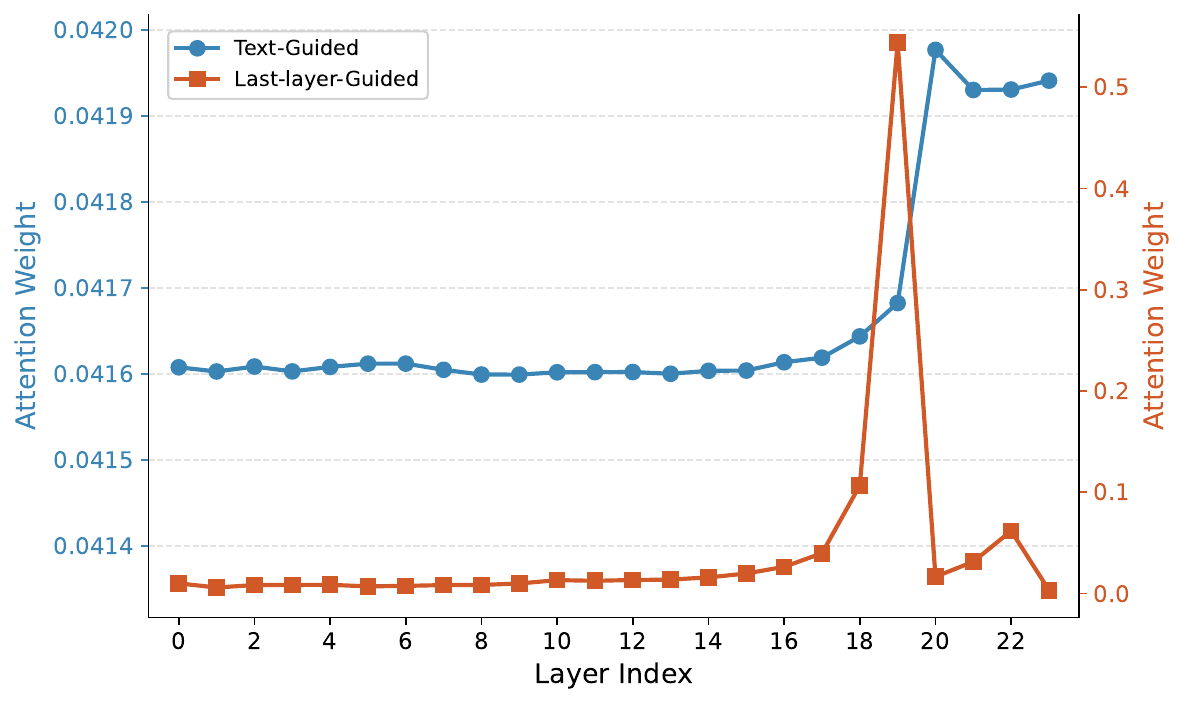} 
        \footnotesize (a) Biased Layer Preference in Multi-layer Fusion
        \label{fig:query-attn}
    \end{subfigure}
    \hfill
    \begin{subfigure}{0.42\linewidth}
        \centering
        \includegraphics[width=\linewidth,height=4.2cm,keepaspectratio]{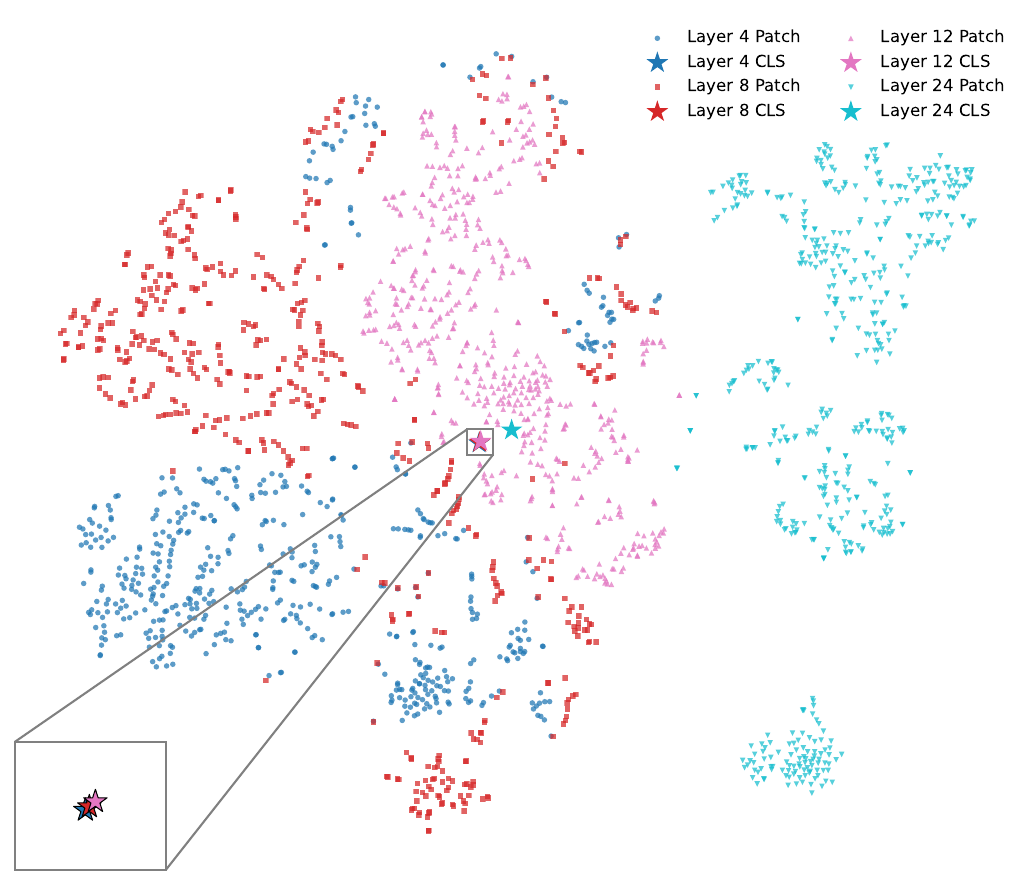}
        \footnotesize (b) t-SNE visualization across layers
        \label{fig:different_layer}
    \end{subfigure}
    \caption{
    Layer-wise visualization of attention patterns and feature embedding distributions in vision encoder.
    (a) ``Text-Guided'' fusion uses textual representations as queries to select visual features from different layers, while ``Last-layer-Guide'' fusion employs final-layer visual representations as queries. Attention analysis reveals both ``Text-Guided'' and ``Last-layer-Guided'' visual feature aggregation consistently prioritize deeper encoder layers over shallow ones.
    (b) t-SNE visualization of visual features from different layers of the vision encoder. 
    Features from shallow and intermediate layers exhibit distinct distributions compared to deeper layers, indicating significant semantic misalignment across layers.}
    \label{fig:analysis}
\end{figure}

Given their favorable efficiency-effectiveness trade-offs, aggregating multi-layer representations has emerged as a widely-embraced scheme for visual enhancement in MLLMs. Despite its promising performance, it still face several limitations. In what follows, we empirically reveal by two key observations:
\textbf{(1) Under-exploration of Hierarchy Visual Feature.} As shown in Fig.~\ref{fig:analysis}(a), although shallow and intermediate features are theoretically modeled for multi-grained understanding in multi-layer fusion scheme, attention analysis demonstrates that they are often  insufficiently leveraged when relying on implicit, static, data-driven feature aggregation. In other words, both text-guided and last-layer-guided aggregation mechanisms exhibit a consistent bias toward deeper-layer representations. This suggests that shallow-layer features—rich in fine-grained spatial details—are not effectively integrated into the final representation, limiting the model's capacity to fully exploit hierarchical visual cues across the encoder depth. \textbf{(2) Hierarchical Vision-Language Misalignment.} For a well-trained MLLM, shallow and intermediate visual features often exhibit significant misalignment with the semantic embedding space of the LLM. As evidenced in Fig.~\ref{fig:analysis}(b), features extracted from the final layer form a tight cluster that is well-separated from both shallow and intermediate layer representations. This limitation stems from the fact that the vision-text projector is typically optimized exclusively using final-layer visual features and textual embeddings. Consequently, shallow and intermediate features lack an effective projection pathway to map into the LLM's space, leading to substantial distributional differences. Therefore, rather than costly full retraining, a model-agnostic, hierarchy-aware adaptation strategy with minimal training overhead is indispensable to exploit underutilized features and mitigate distributional misalignment.

To address these limitations, we introduce a \textbf{Me}mory-Aug\textbf{m}ented \textbf{A}dapter \textbf{(Mema)} that facilitates dynamic, explicit cross-layer interaction while rectifying hierarchical misalignment within the vision encoder. The overall framework is illustrated in Fig.~\ref{fig:comp}(c). Specifically, Mema introduces a stateful memory that accumulates hierarchical visual representations across layers, with its evolution conditioned on both query embeddings and step-wise visual features. At each step, the memory state is updated by integrating current visual-semantic features with prior memory contexts. A portion of this memory is selectively injected into the main encoding branch via a feedback mechanism to balance the global state and layer-wise preference, which effectively achieves comprehensive cross-layer hierarchy alignment. Beyond that, this design allows shallow-layer features to be propagated not only sequentially but also directly through a bypass pathway, mitigating the attenuation of fine-grained visual cues.

Moreover, textual queries often contain generic words (e.g., ``how many'', ``what''), which may cause the memory state to drift toward irrelevant semantics during accumulation. To mitigate this, we introduce an auxiliary alignment objective that supervises the evolution of memory, which effectively accelerates convergence towards task-relevant intentions.
Without bells and whistles, our designed lightweight adapter can be integrated seamlessly into pretrained vision encoders, which enables adaptive visual feature refinement without requiring substantial modification to the visual backbone or training from scratch. Furthermore, it requires only a minimal number of additional parameters to adapt to fine-grained visual cues, ensuring parameter-efficient deployment.

We summarize our contributions as follows:

\begin{itemize}

\item We propose Mema, a dynamic memory-augmented adapter that enables hierarchical visual reservation and alignment within the vision encoder.

\item We devise two lightweight modules that effectively preserves long-term memory with hierarchical consistency and guide memory evolution aligned with task-specific semantics.

\item Mema is a lightweight and plug-and-play framework that enables efficient adaptation within pretrained vision encoders without modifying the backbone or the language model, and generalizes effectively across diverse architectures on a series of multimodal benchmarks.

\end{itemize}

\section{Related Work}
\label{related work}

\textbf{\emph{Multimodal Large Language Models.}} 
Multimodal Large Language Models (MLLMs) have emerged as a dominant paradigm for vision-language understanding, enabling joint reasoning over images and text. Early representative works, such as Flamingo~\cite{flamingo} and PaLI~\cite{pail}, demonstrated the effectiveness of large-scale vision-language pretraining. Subsequent approaches, including BLIP-2~\cite{blip2} and LLaVA~\cite{llava}, efficiently align visual features with language models through various learnable connectors without requiring full model retraining. More recent models, such as InstructBLIP~\cite{instructblip}, Kosmos-2~\cite{kosmos-2}, and Qwen2-VL~\cite{qwen2-vl}, further enhance multimodal capabilities by strengthening instruction-following abilities and fine-grained visual grounding through data scaling.

Despite their promising performance, these existing approaches generally rely on fixed visual representations extracted from pretrained vision encoders, constraining their capacity to adapt to task-specific requirements during visual encoding.

\noindent\textbf{\emph{Enhancing Visual Representations in MLLMs.}} 
To improve the task adaptation of visual representation in multimodal large language models, prior work mainly explores from three directions. One line of research enhances visual perception by increasing the capacity of vision encoders. For example, PaLI~\cite{pail} scales the vision backbone to a billion-parameter Vision Transformer~\cite{vit}. InternVL~\cite{internvl} further extends this paradigm by scaling the vision encoder to multi-billion parameter levels.
PaLM-E~\cite{palm-e} integrates a large-scale Vision Transformer with a 540B-parameter language model, enabling tightly coupled multimodal reasoning across modalities. 
Another strand of research improves visual perception by integrating multiple vision encoders, leveraging their distinct visual biases for collaborative visual encoding. For instance, MMVP~\cite{mmvp} directly combines features from multiple pretrained vision models (e.g., CLIP~\cite{clip} and DINOv2~\cite{dinov2}) to enrich visual representations. In contrast, MGM~\cite{mgm} leverages high-resolution visual representations to enhance the low-resolution visual branch via cross-attention. Similarly, LEO~\cite{leo} adopts a dual-branch architecture with interleaved token fusion to exploit complementary representations across encoders.

To improve the encoding efficiency, a series of concurrent works focus on hierarchical feature exploitation within individual vision encoders. Dense Connector~\cite{dc} pioneers concatenation-based multi-layer fusion, while MMFuser~\cite{mmfuser} introduces cross-layer querying from high-level to shallow features. More recently, IGF~\cite{instruction-guided} and TGIF~\cite{tgif} advance this paradigm by incorporating textual queries, enabling text-aware dynamic aggregation of multi-layer visual features. However, these approaches either rely on scaling or static feature aggregation after visual encoding, which lacks of dynamic and sufficient cross-layer feature interaction and alignment.

\noindent\textbf{\emph{Cross-layer Interaction in Deep Networks.}} 
Prior work has explored modeling inter-layer dependencies to enhance representation learning. For instance, DIANet~\cite{dia} employs a parameter-sharing LSTM module across network depth to enable information exchange among different layers. MRLA~\cite{mrla} sends a query representation from the current layer to all previous layers, retrieving query-related information from different levels of receptive fields, which effectively leverages cross-layer dependencies. In particular, DLA~\cite{dsu} unlocks context feature extraction across layers through a dual-path feature refinement scheme. Despite their advances, most existing approaches are designed for convolution-based vision models, and adapting them to large-scale vision-language understanding remains an open problem.
\begin{figure*}
    \centering
    \includegraphics[width=0.95\linewidth]{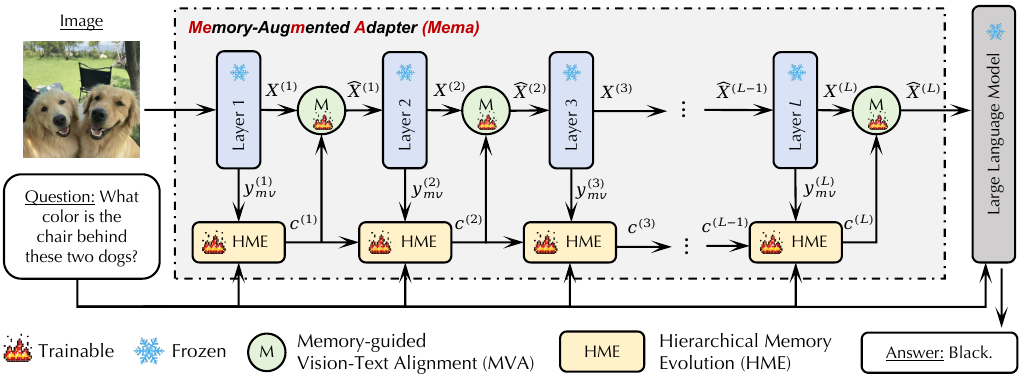}
    \caption{
Overview of Mema. Mema maintains a stateful memory that accumulates hierarchical visual representations via HME, and injects it back into token representations through MVA. 
}
    \label{fig:scvm}
\end{figure*}

\begin{figure}
    \centering
    \includegraphics[width=\linewidth]{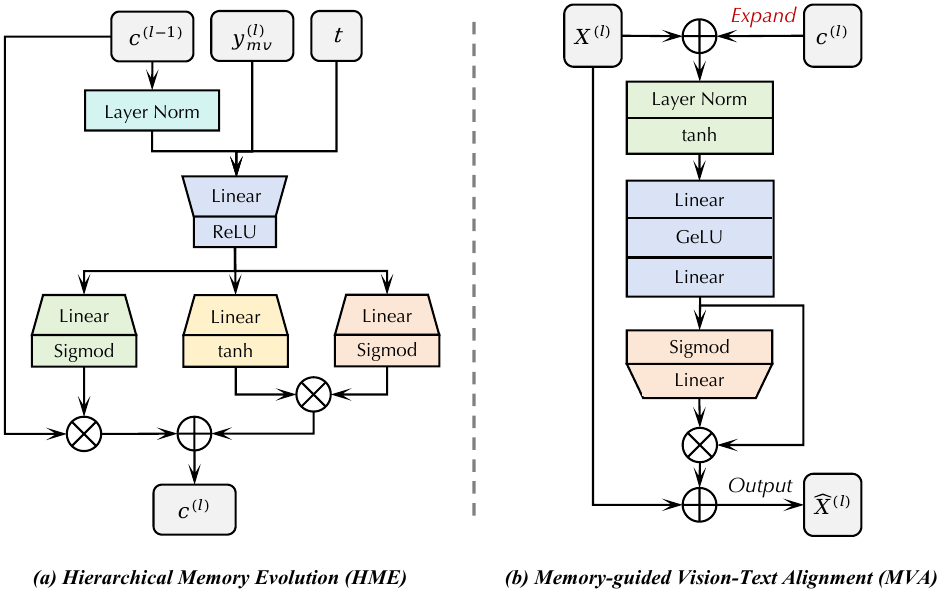}
    \caption{
Illustration of (a) HME and (b) MVA.
}
    \label{fig:HME-MVA}
\end{figure}

\section{Method}

In this section, we present Mema within the vision encoder, serving as the bridge for dynamic multi-layer feature exploitation and hierarchy visual alignment. 
We first provide an overview in Section~\ref{sec:overview}. Then, we elaborate two main components: Hierarchical Memory Evolution (HME) in Section~\ref{sec:HME} and Memory-Guided Refinement (MVA) in Section~\ref{sec:MVA}. 
Section~\ref{sec:MVA} also introduces an auxiliary semantic alignment objective to ensure task-aligned memory representations.

\subsection{Overview}
\label{sec:overview}
Given an image, we denote the $\mathbf{X}^{(l)} \in \mathbb{R}^{N \times D}$ as the visual feature extracted from the $l$-th layer, where $N$ is the number of tokens and $D$ is the hidden dimension. 
We first construct a layer-specific summary token $\mathbf{y}_{\mathrm{mv}}^{(l)} \in \mathbb{R}^{D}$ from $\mathbf{X}^{(l)}$, and maintain a dynamic stateful memory across layers. 
At each layer, the memory state is updated using the current summary representation, a global textual representation $\mathbf{t} \in \mathbb{R}^{D}$ corresponds to the input query, and the previous memory state,
\begin{align}
\mathbf{c}^{(l)} = \mathrm{HME}(\mathbf{y}_{\mathrm{mv}}^{(l)}, \mathbf{t}, \mathbf{c}^{(l-1)}).
\end{align}

The updated memory state $\mathbf{c}^{(l)}$ is then broadcast and fed back to refine the visual representation $\mathbf{X}^{(l)}$ in a token-wise manner, updating the representations as follows,
\begin{align}
\hat{\mathbf{X}}^{(l)} = \mathrm{MVA}(\mathbf{X}^{(l)}, \mathbf{c}^{(l)}).
\end{align}
In what follows, we elaborate the details of two designed components, HME and MVA.

\subsection{Hierarchical Memory Evolution (HME)}
\label{sec:HME}
Our memory update mechanism draws inspiration from gated recurrent units, such as LSTMs~\cite{lstm}, which maintain a hidden state to capture global context across sequential input steps. In contrast, we introduce HME to iteratively evolve the memory state across hierarchical transformer layers, as depicted in Fig.~\ref{fig:HME-MVA}(a).

At each layer, the memory is updated using the current visual features, textual semantics and previous memory prior, allowing visual cues from different layers to be progressively incorporated into the memory state. Specifically, visual features are summarized via a multi-view aggregation strategy,
\begin{align}
\mathbf{y}_{\mathrm{mv}}^{(l)} =
\mathrm{MLP} \left(
\mathrm{Avg}(\mathbf{X}^{(l)}) ||
\mathrm{Max}(\mathbf{X}^{(l)}) ||
\mathbf{x}_{\mathrm{CLS}}^{(l)}
\right)
\in \mathbb{R}^{D},
\end{align}
where average pooling captures global context, max pooling highlights salient regions, and the CLS token provides semantic summary. These three complementary operations ensure comprehensive visual summarization. Additionally, we incorporate the global textual representation $\mathbf{t} \in \mathbb{R}^{D}$, which is obtained by average pooling over input query embeddings from the LLM tokenizer. This serves as query-aware guidance to keep the memory state aligned with the user intent. Along with the previous memory state $\mathbf{c}^{(l-1)}$, we finally obtain a new representation as follows,
\begin{align}
\mathbf{s}^{(l)} = \mathrm{ReLU}\left(\mathbf{W}_s \left(\mathrm{LN}(\mathbf{c}^{(l-1)})|| \mathbf{y}_{\mathrm{mv}}^{(l)}|| \mathbf{t}\right)\right),
\end{align}
where $\mathbf{W}_s \in \mathbb{R}^{d_r \times 3D}$ is the projection matrix, which projects the concatenated input into low-dimension representation with a reduced dimension $d_r = \frac{D}{r}$, and $\mathrm{LN}(\cdot)$ denotes the layer normalization. Based on $\mathbf{s}^{(l)}$, the generated memory content and gating signals in current step are computed as,
\begin{equation}
\left\{
\begin{aligned}
\tilde{\mathbf{c}}^{(l)} &= \tanh(\mathbf{W}_c \mathbf{s}^{(l)} + \mathbf{b}_c), \\
\mathbf{i}^{(l)} &= \sigma(\mathbf{W}_i \mathbf{s}^{(l)} + \mathbf{b}_i), \\
\mathbf{f}^{(l)} &= \sigma(\mathbf{W}_f \mathbf{s}^{(l)} + \mathbf{b}_f),
\end{aligned}
\right.
\end{equation}
where $\mathbf{W}_c, \mathbf{W}_i, \mathbf{W}_f \in \mathbb{R}^{D \times d_r}$ are learnable projection matrices and $\mathbf{b}_c, \mathbf{b}_i, \mathbf{b}_f \in \mathbb{R}^{D}$ are bias terms. The final memory state is weighted by,
\begin{align}
\mathbf{c}^{(l)} = \mathbf{f}^{(l)} \odot \mathbf{c}^{(l-1)} + \mathbf{i}^{(l)} \odot \tilde{\mathbf{c}}^{(l)}.
\end{align}
This strategy enables the memory to encapsulate hierarchical visual cues from early layers. Subsequently, the obtained memory state $\mathbf{c}^{(l)}$ can be either transmitted to the main processing branch (i.e., the transformer pathway) or routed through a bypass pathway (i.e., directly propagated to the next layer iteration). This dual-pathway design effectively mitigates the attenuation of hierarchy visual feature throughout the integration process.

\subsection{Memory-guided Vision-Text Alignment (MVA)}
\label{sec:MVA}

To further enable shallow and intermediate visual features to align with the semantic embedding space of the LLM without modifying the pretrained projector, we introduce the MVA, as depicted in Fig.~\ref{fig:HME-MVA}(b). Specifically, let $\mathbf{X}^{(l)} \in \mathbb{R}^{N \times D}$ denotes the visual features at layer $l$, and $\mathbf{c}^{(l)} \in \mathbb{R}^{D}$ is the memory state generated from HME module. 
We first broadcast $\mathbf{c}^{(l)}$ and combine it with the visual features in the current layer to form a memory-guided representation,
\begin{align}
\mathbf{H}^{(l)} = \mathrm{LN}(\mathbf{X}^{(l)} + \mathbf{c}^{(l)}).
\end{align}
Based on $\mathbf{H}^{(l)}$, we compute a shared representation bridging global and layer-specific features, along with a token-wise gating factor for adaptive feature propagation,
\begin{align}
\boldsymbol{\Delta}^{(l)} &= \tanh(\mathrm{MLP}(\mathbf{H}^{(l)})), \\
\boldsymbol{\alpha}^{(l)} &= \sigma(\mathbf{W}_h \mathbf{H}^{(l)}),
\end{align}

where $\boldsymbol{\Delta}^{(l)} \in \mathbb{R}^{N \times D}$ represents the candidate residual update, while $\boldsymbol{\alpha}^{(l)} \in \mathbb{R}^{N \times D}$ controls its update magnitude, preventing overly large variation that may disrupt the distribution of original representations.
The updated token representation is,
\begin{align}
\hat{\mathbf{X}}^{(l)} = \mathbf{X}^{(l)} + \boldsymbol{\alpha}^{(l)} \odot \boldsymbol{\Delta}^{(l)},
\end{align}
where $\odot$ denotes element-wise multiplication. In a deeper analysis, this mechanism intrinsically functions as a filtering module, aligning previous hierarchical visual memory with the distribution of current visual features. Consequently, it progressively bridges the semantic gap between hierarchical representations and the final-layer visual distribution (i.e., LLM's semantic embedding space) with the layer-by-layer refinement.

To further mitigate interference from ambiguous semantics conveyed in textual queries, we introduce an auxiliary alignment objective that explicitly supervises memory evolution. Let $\mathbf{c}^{(L)} \in \mathbb{R}^{D}$ denote the final-layer memory state, and we first project it into the language space,
\begin{align}
\mathbf{c}_{\mathrm{final}} = \mathrm{MLP}(\mathbf{c}^{(L)}),
\end{align}
where $\mathbf{c}_{\mathrm{final}} \in \mathbb{R}^{D_{\mathrm{llm}}}$.
The alignment loss is defined as,
\begin{align}
\mathcal{L}_{\mathrm{align}} = 1 - \cos(\mathbf{c}_{\mathrm{final}}, \mathbf{a}),
\end{align}
where $\mathbf{a} \in \mathbb{R}^{D_{\mathrm{llm}}}$ denotes the average pooling representation from $\mathbf{t}_{\mathrm{ans}} \in \mathbb{R}^{T_a \times D_{\mathrm{llm}}}$ answer token embedded by LLM tokenizer, $T_{a}$ is the number of answer words. The overall training objective is,
\begin{align}
\mathcal{L} = \mathcal{L}_{\mathrm{MLLM}} + \lambda \mathcal{L}_{\mathrm{align}},
\end{align}
where $\lambda$ is a hyperparameter balancing the autoregressive loss and the alignment objective. Through gradient backpropagation, this supervision signal propagates to intermediate memory states, guiding the overall memory trajectory toward task-relevant semantics during evolution. 
\section{Experiments}

\begin{table*}[t]
\centering
\setlength{\tabcolsep}{2pt}  
\small
\caption{Plug-and-play transferability of Mema across different multimodal backbones. 
Mema is trained once on LLaVA-v1.5~\cite{llava} and directly inserted into other models with compatible hidden dimensions. 
$^*$ indicates a CLIP ViT-L/14-336~\cite{clip} vision tower further fine-tuned on ShareGPT4V~\cite{sharegpt4v} data.}
\label{tab:main_results}
\begin{tabular}{lccccccccccc}
\toprule
Model & Vision Encoder & LLM & TextVQA & GQA & MME$^{P}$ & MME$^{C}$ & SQA & POPE & MMB & MMB$^{CN}$ & SEED$^{I}$ \\
\midrule

\multicolumn{12}{c}{\textbf{Mema trained on LLaVA-v1.5-7B}} \\
\midrule

LLaVA-v1.5 
& \multirow{2}{*}{CLIP ViT-L/14-336} 
& \multirow{2}{*}{Vicuna-7B}
& 58.2 & \textbf{62.0} & 1510.7 & 355.7 & 66.8 & 85.9 & 64.3 & 58.3 & 66.0 \\
+ Mema
& & 
& 58.2 & 61.8 & \textbf{1520.6} & \textbf{372.2} & \textbf{70.1} & \textbf{86.7} & \textbf{64.7} & 58.3 & 66.0 \\

\midrule

LLaVA-v1.5 
& \multirow{2}{*}{OpenCLIP ViT-L/14} 
& \multirow{2}{*}{Vicuna-7B}
& 53.9 & 58.5 & 1332.0 & \textbf{321.1} & \textbf{71.0} & 84.1 & 60.8 & 52.3 & 64.2 \\
+ Mema
& & 
& \textbf{54.1} & \textbf{60.9} & \textbf{1418.1} & 311.0 & 70.8 & \textbf{85.3} & \textbf{65.5} & \textbf{57.8} & \textbf{64.4} \\

\midrule

MGM 
& \multirow{2}{*}{CLIP ViT-L/14-336 + ConvNeXt-L} 
& \multirow{2}{*}{Vicuna-7B}
& 65.2 & 62.6 & 1523.0 & 316.0 & 71.5 & \textbf{87.0} & \textbf{69.3} & 57.4 & 69.4 \\
+ Mema
& & 
& 65.2 & 62.6 & \textbf{1541.6} & \textbf{326.4} & \textbf{71.6} & 86.7 & 68.8 & \textbf{58.2} & 69.4 \\

\midrule

ShareGPT4V 
& \multirow{2}{*}{CLIP ViT-L/14-336$^*$} 
& \multirow{2}{*}{Vicuna-7B}
& 60.4 & 63.3 & 1567.4 & \textbf{376.4} & 68.4 & 86.8 & 68.8 & \textbf{62.2} & 69.7 \\
+ Mema
& & 
& 60.4 & 63.3 & \textbf{1573.3} & 356.1 & \textbf{71.0} & 86.8 & \textbf{68.9} & 61.5 & 69.7 \\

\midrule
\multicolumn{12}{c}{\textbf{Mema trained on LLaVA-v1.5-13B}} \\
\midrule

LLaVA-v1.5 
& \multirow{2}{*}{CLIP ViT-L/14-336} 
& \multirow{2}{*}{Vicuna-13B}
& \textbf{61.3} & 63.3 & 1531.3 & 296.1 & 71.6 & 85.9 & 67.7 & 63.6 & 61.6 \\
+ Mema
& & 
& 61.0 & \textbf{63.4} & \textbf{1534.1} & \textbf{297.5} & \textbf{75.0} & \textbf{87.2} & \textbf{68.6} & \textbf{63.8} & \textbf{68.4} \\

\midrule

MGM 
& \multirow{2}{*}{CLIP ViT-L/14-336 + ConvNeXt-L} 
& \multirow{2}{*}{Vicuna-13B}
& 65.9 & 63.3 & 1564.0 & 321.0 & 76.1 & 86.4 & 68.5 & 63.1 & 70.4 \\
+ Mema
& & 
& \textbf{66.1} & \textbf{63.4} & \textbf{1583.1} & \textbf{333.6} & 76.1 & \textbf{86.6} & \textbf{68.9} & 63.1 & \textbf{71.1} \\

\midrule

ShareGPT4V 
& \multirow{2}{*}{CLIP ViT-L/14-336$^*$} 
& \multirow{2}{*}{Vicuna-13B}
& 62.2 & 64.8 & \textbf{1618.7} & 303.2 & 71.2 & 87.3 & 68.5 & 63.7 & 70.8 \\
+ Mema
& & 
& \textbf{62.5} & 64.8 & 1597.8 & \textbf{310.0} & \textbf{73.7} & \textbf{87.4} & \textbf{69.2} & \textbf{64.1} & \textbf{71.0} \\

\bottomrule
\end{tabular}
\end{table*}

\begin{table}[t]
\centering
\small
\setlength{\tabcolsep}{4pt} 
\caption{
Parameter and training cost comparison of Mema on LLaVA-1.5-7B~\cite{llava}. 
``Trainable'' denotes the number of trainable parameters, ``Total'' refers to the overall model size, 
``FLOPs'' indicates the forward computation per batch, and ``Data'' denotes the number of training samples. 
``+ Full FT'' denotes fine-tuning both the LLM and the multimodal projector, while ``+ Proj Only'' denotes fine-tuning only the multimodal projector. 
Total training FLOPs can be computed as FLOPs $\times$ Data.
}
\label{tab:parameter}
\begin{tabular}{lcccc} 
\toprule 
Method & Trainable (M) & Total (B) & FLOPs (T) & Data (K) \\ 
\midrule 
+ Full FT & 6628.8 & 6.932 & 457.5 & 665 \\ 
+ Proj Only & 21.0 & 6.932 & 307.4 & 558 \\ 
+ Mema & 11.0 & 6.943 & 315.9 & 20 \\ 
\bottomrule 
\end{tabular}
\end{table}

\begin{table*}[t]
\centering
\setlength{\tabcolsep}{3pt}  
\small
\caption{
Generalization of Mema across different vision encoders with heterogeneous feature dimensions. 
}
\label{tab:encoder_results}
\begin{tabular}{lcccccccccccc}
\toprule
Model & Vision Encoder & LLM & TextVQA & GQA & MMMU & VQAv2 & MME$^{P}$ & SQA & POPE & MMB & MM-Vet \\
\midrule

Mipha 
& \multirow{2}{*}{SigLIP-SO400M/14-384} 
& \multirow{2}{*}{Phi-1.5-1.3B}
& 45.6 & \textbf{62.7} & \textbf{29.9} & 77.5 & 1203.1 & 58.3 & 86.9 & 57.7 & 23.5 \\
+ Mema  
& & 
& \textbf{49.3} & 62.5 & 29.3 & \textbf{79.2} & \textbf{1285.5} & \textbf{61.2} & 86.9 & \textbf{60.7} & \textbf{26.7} \\

\midrule

Mipha 
& \multirow{2}{*}{SigLIP-SO400M/14-384} 
& \multirow{2}{*}{Phi-2-2.7B}
& \textbf{56.6} & 63.9 & 25.2 & 81.3 & 1488.9 & 70.9 & 86.7 & 69.7 & 31.3 \\
+ Mema  
& & 
& 56.2 & \textbf{64.0} & \textbf{26.8} & 81.3 & \textbf{1494.7} & \textbf{74.6} & \textbf{87.8} & \textbf{70.0} & \textbf{33.1} \\

\midrule

LLaVA-v1.5
& \multirow{2}{*}{SigLIP-B/16-224} 
& \multirow{2}{*}{Vicuna-7B}
& \textbf{53.4} & 61.0 & 33.8 & \textbf{76.0} & 1427.4 & 70.2 & \textbf{84.2} & \textbf{63.1} & 26.7 \\
+ Mema  
& & 
& 52.4 & 61.0 & \textbf{34.4} & 75.0 & \textbf{1451.6} & \textbf{70.3} & 83.9 & 62.0 & \textbf{27.1} \\

\bottomrule
\end{tabular}
\end{table*}

\begin{table}[t]
\centering
\small
\setlength{\tabcolsep}{2pt}  
\caption{
Comparison with representative multi-layer visual fusion methods.
}
\label{tab:sotamodel}
\begin{tabular}{lccccc}
\toprule
Model  & MME$^{P}$ & SQA & MMB & POPE & Avg \\
\midrule
LLaVA-1.5-7B & 1510.7 & 66.8 & 64.3 & 85.9 & 73.1 \\
Dense Connector~\cite{dc} & -- & 69.5 & 66.8 & 86.6 & -- \\
MMFuser~\cite{mmfuser} & 1479.7 & 68.7 & \textbf{67.5} & 86.3 & 74.1 \\
Instruction-Guided Fusion~\cite{instruction-guided} & 1519.8 & \textbf{70.2} & 66.9 & 87.8 & \textbf{75.2} \\
TGIF~\cite{tgif} & -- & 70.1 & 66.4 & \textbf{87.9} & -- \\
\midrule
+ Mema & \textbf{1520.6} & 70.1 & 64.7 & 86.7 & 74.4 \\
\bottomrule
\end{tabular}
\end{table}

\subsection{Implementation Details}

\textbf{\emph{Architecture.}} 
We implement Mema based on LLaVA-v1.5-7B~\cite{llava} with a CLIP ViT-L/14-336~\cite{clip} encoder. 
HME and MVA are inserted into the vision encoder, and both the vision backbone and LLM remain frozen.

To encourages conservative behavior during early training and enhance gradient stability, we adopt the following setting: For HME, the forget-gate bias is set to $1.0$, while other biases are initialized to zero. 
The reduction ratio is set to $r=4$.
For MVA, the modulation MLP is initialized with near-zero weights, and the gate bias is set to $-2.2$ (i.e., $\sigma(b)\approx0.1$).

\noindent\textbf{\emph{Backbone Variants.}} 
We also evaluate the effect of our Mema on different vision encoders and LLMs. 
Vision encoders include CLIP~\cite{clip} (ViT-L/14-336), in both its original pretrained form and a variant further fine-tuned on ShareGPT4V~\cite{sharegpt4v}, as well as OpenCLIP~\cite{openclip} (ViT-L/14) and SigLIP~\cite{siglip} (e.g., SO400M/14-384 and B/16-224). 
We also consider multi-backbone architectures such as MGM~\cite{mgm}.
For MGM, Mema is applied only to the CLIP branch, leaving the ConvNeXT-based~\cite{convnext} branch unchanged. 
Language models include Vicuna-7B/13B and the Phi series~\cite{phi-1.5,phi-2}.

\noindent\textbf{\emph{Training Setup.}} 
We train with DeepSpeed ZeRO-2 on two A100 GPUs using bf16 precision. 
For LLaVA-v1.5-7B~\cite{llava}, we use AdamW with a base learning rate of $1\times10^{-4}$ and cosine decay, and set the alignment weight to $\lambda=0.1$. The global batch size is 16. 
For other model variants, hyperparameters are adjusted accordingly. 
Additional implementation details are provided in Appendix.

\noindent\textbf{\emph{Dataset.}} 
We randomly sample 20K instances from LLaVA-Instruct-665K~\cite{llava}  to train only the Mema, while keeping the backbone frozen.

\noindent\textbf{\emph{Benchmark.}} 
We evaluate on standard multimodal benchmarks including VQAv2~\cite{vqav2}, GQA~\cite{gqa}, TextVQA~\cite{textvqa}, ScienceQA~\cite{sqa}, MME~\cite{mme}, POPE~\cite{pope}, MMBench~\cite{mmb}, MMBench$^{CN}$~\cite{mmb}, MMMU~\cite{mmmu}, SEED$^{I}$~\cite{seed}, and MM-Vet~\cite{mmv}.

\subsection{Main Results}

\noindent\textbf{\emph{Plug-and-Play Transfer Across Compatible Vision Encoders.}} 
Table~\ref{tab:main_results} reports the plug-and-play performance gain of Mema on different vision encoders. We train Mema once on LLaVA-v1.5 backbone and directly apply it to other models without modifying the architecture or retraining the LLM. 
Notably, Mema introduces only 11.0M trainable parameters, compared to 6628.8M in full fine-tuning and 21.0M when training only the multimodal projector (Table~\ref{tab:parameter}). 
Combined with its significantly reduced training data and computational cost, this highlights the strong parameter and compute efficiency of Mema while maintaining consistent performance gains. 

In the 7B setting, Mema improves the original LLaVA-v1.5 backbone, with gains on MME$^{P}$ (+9.9), MME$^{C}$ (+16.5), and SQA (+3.3), and generalizes effectively to different vision encoders. 
For example, when applied to OpenCLIP, it yields consistent gains across multiple benchmarks, including MME$^{P}$ (+86.1) and MMB (+4.7). These gains are more pronounced when the underlying visual representations are weaker or less aligned. 
Mema also generalizes to more complex architectures such as MGM, improving MME$^{P}$ (+18.6), MME$^{C}$ (+10.4), as well as downstream benchmarks such as SQA and MMB. 
Even on ShareGPT4V, where the vision encoder has already been adapted, Mema still provides consistent improvements, e.g., SQA (+2.6).

A similar trend is observed in the 13B setting. 
Mema consistently improves performance across different architectures, including LLaVA-v1.5, MGM, and ShareGPT4V, on multiple benchmarks such as MME$^{P}$, MME$^{C}$, SQA, MMB, and SEED$^{I}$. 
For instance, it achieves gains on SQA (+3.4) and MMB (+0.9) for LLaVA-v1.5-13B, and improves MME$^{P}$ (+19.1) and MME$^{C}$ (+12.6) on MGM. 
Even when the baseline performance is already strong, Mema continues to yield improvements, indicating its effectiveness and scalability across model sizes.

\begin{figure}[t]
    \centering
    \begin{subfigure}{0.46\linewidth}
        \centering
        \includegraphics[width=\linewidth,height=4.4cm,keepaspectratio]{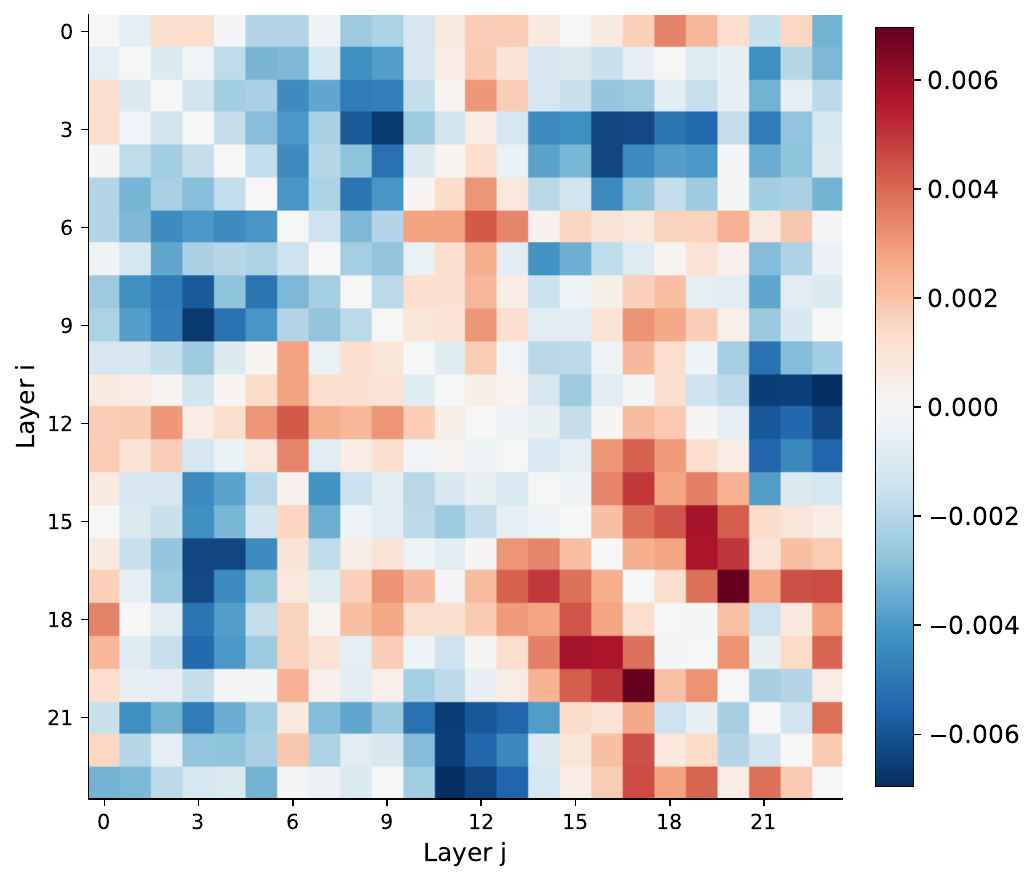}
        \footnotesize (a) Cross-layer similarity difference
        \label{fig:cos-similarity}
    \end{subfigure}
    \hfill
    \begin{subfigure}{0.50\linewidth}
        \centering
        \includegraphics[width=\linewidth,height=4.4cm,keepaspectratio]{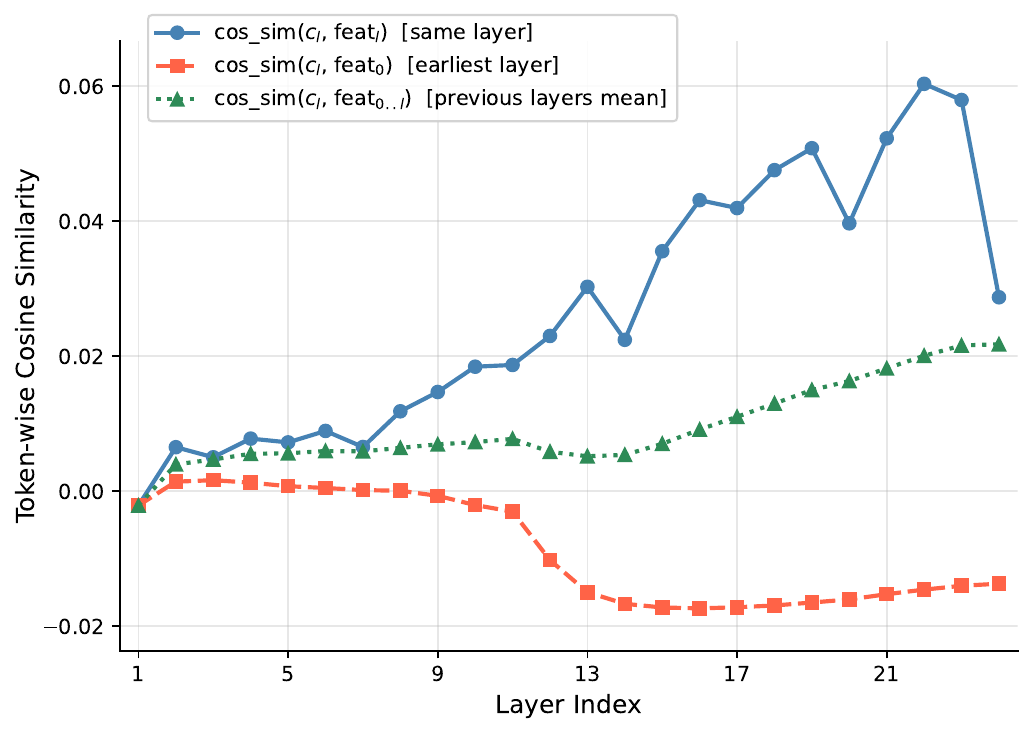}
        \footnotesize (b) Memory dynamics and alignment across layers
        \label{fig:memory_effectiveness}
    \end{subfigure}
    
    \caption{
    Visualization analysis of behavior of our method. (a) Cross-layer similarity difference bewteen Mema and vanilla LLaVA-1.5-7B across vision layers. Red indicates increased similarity under Mema, while blue indicates decreased similarity. Mema reshapes the interaction pattern across depth.
    (b) The cosine similarity between the memory state and the current-layer features, the earliest-layer features, and the mean of features from all preceding layers, indicating progressive alignment and effective cross-layer information aggregation.
    }
    \label{fig:memory_analysis}
\end{figure}

\begin{table}[t]
\centering
\setlength{\tabcolsep}{1pt}  
\small
\caption{Effect of progressive cross-layer interaction. 
MVA is applied at the last layer (Last layer), selected layer subsets, odd-numbered layers (Odd Layers), or all layers (Ours).}
\label{tab:interaction_effect}
\begin{tabular}{lccccc}
\toprule
Method & TextVQA & MME$^{P}$ & MME$^{C}$ & MMB & Avg  \\
\midrule
LLaVA-v1.5-7B & 58.20 & 1510.7 & 355.7 & 64.30 & 60.63  \\
Last layer & 58.00 & 1517.9 & 353.2 & 64.30 & 60.60  \\
Layers \{8,16,24\} & 58.00 & 1516.2 & 350.4 & 64.50 & 60.53  \\
Layers \{4,8,12,16,20,24\} & 58.00 & 1514.5 & 343.6 & 64.40 & 60.28  \\
Odd Layers & 58.20 & 1500.5 & 357.5 & \textbf{65.00} & 60.73  \\
\midrule
All Layers (Ours) & 58.20 & \textbf{1520.6} & \textbf{372.2} & 64.70 & \textbf{61.35}  \\
\bottomrule
\end{tabular}
\end{table}

\noindent\textbf{\emph{Adaptation to Heterogeneous Vision Encoders.}} 
Table~\ref{tab:encoder_results} presents the performance of Mema on heterogeneous vision encoders, including SigLIP-based~\cite{siglip} architectures and lightweight MLLMs.

Mema consistently improves performance across all evaluated settings. 
On Mipha~\cite{mipha} with SigLIP-SO400M~\cite{siglip}, it achieves gains across multiple benchmarks, including TextVQA, VQAv2, MME$^{P}$, SQA, MMB, and MM-Vet. 
These improvements remain consistent when scaling the LLM from 1.3B to 2.7B, indicating that Mema generalizes across different model sizes. 
Similarly, on LLaVA with a SigLIP-B/16-224~\cite{siglip} encoder, Mema yields consistent gains across benchmarks.

These results demonstrate that Mema generalizes effectively across diverse visual backbones and model scales through lightweight adaptation, learning a transferable cross-layer interaction pattern while maintaining compatibility with pretrained models.

\begin{table}[t]
\centering
\small
\setlength{\tabcolsep}{3pt}  
\caption{Ablation study of different components in Mema.}
\label{tab:ablation}
\begin{tabular}{lccccc}
\toprule
Method & TextVQA & MME$^{P}$ & MME$^{C}$ & MMB & Avg \\
\midrule
LLaVA-v1.5-7B & 58.2 & 1510.7 & 355.7 & 64.3 & 60.63 \\
\midrule
+ Mema w/o Text & \textbf{58.4} & 1489.4 & 371.1 & 64.2 & 60.88 \\
+ Mema w/o HME & 58.2 & 1510.0 & 355.7 & 64.6 & 60.70 \\
+ Mema w/o MVA & 29.9 & 1513.0 & 355.7 & 20.5 & 42.65 \\
\midrule
+ Mema & 58.2 & \textbf{1520.6} & \textbf{372.2} & \textbf{64.7} & \textbf{61.35} \\
\bottomrule
\end{tabular}
\end{table}

\definecolor{baselinegray}{gray}{0.9}
\definecolor{oursblue}{RGB}{220,230,255}
\begin{table}[t]
\centering
\setlength{\tabcolsep}{3pt}  
\small
\caption{
Sensitivity analysis of Mema with respect to the reduction ratio $r$ and the alignment weight $\lambda$. 
The \colorbox{oursblue}{blue row} denotes our default setting.
We report performance across various benchmarks, with the best results in bold.
}
\begin{tabular}{cccccccc}
\toprule
$r$ & $\lambda$ & MME$^{P}$ & MME$^{C}$ & SQA & POPE & MMB & Avg \\
\midrule
\multicolumn{2}{c}{LLaVA-v1.5-7B} & 1,510.7 & 355.7 & 66.8 & 85.9 & 64.3 & 67.4 \\
\midrule
1 & 0.1 & 1,494.1 & \textbf{382.6} & 70.0 & 86.5 & 64.3 & 68.7 \\
2 & 0.1 & 1,504.0 & 341.1 & \textbf{70.3} & 86.7 & 64.4 & 67.9 \\
4 & 0.05 & 1,484.0 & 357.5 & 70.1 & \textbf{86.9} & 64.8 & 68.1 \\
\rowcolor{oursblue}
4 & 0.1 & \textbf{1,520.6} & 372.2 & 70.1 & 86.7 & 64.7 & \textbf{68.8} \\
4 & 0.15 & 1,497.8	& 357.1 & 70.2 & 86.6 & \textbf{64.9} & 68.2 \\
4 & 0.2 & 1,516.0 & 347.5 & 70.0 & 86.6 & 64.2 & 68.0 \\
8 & 0.1 & 1,504.5 & 350.4 & 70.2 & 86.8 & 64.4 & 68.1 \\
\bottomrule
\end{tabular}
\label{tab:hyperparam}
\end{table}

\noindent\textbf{\emph{Comparison with Multi-Layer Fusion Methods.}} 
Table~\ref{tab:sotamodel} compares Mema with representative approaches that leverage multi-layer visual representations. 

Prior methods, such as Dense Connector~\cite{dc}, MMFuser~\cite{mmfuser}, IGF~\cite{instruction-guided}, and TGIF~\cite{tgif}, typically aggregate features from multiple layers to enhance visual inputs and are often trained jointly with the backbone during early-stage optimization. 

In contrast, Mema requires training only lightweight modules, enabling efficient adaptation with minimal data. 
Despite this reduced training cost, Mema achieves competitive performance across multiple benchmarks, demonstrating that explicit cross-layer interaction provides a more effective and efficient alternative to static multi-layer fusion.

\noindent\textbf{\emph{Cross-Layer Memory Analysis.}}
To understand how the proposed memory mechanism shapes representation evolution, we analyze its behavior across layers. 
As illustrated in Fig.~\ref{fig:memory_analysis}(a), Mema reshapes the interaction pattern across depth, enabling progressive aggregation while preserving meaningful feature diversity. In shallow layers, representations remain relatively distinct, preserving local structures. In intermediate layers, cross-layer information propagation is strengthened, facilitating semantic aggregation. In deeper layers, representations become more globally aligned while maintaining meaningful diversity.
Fig.~\ref{fig:memory_analysis}(b) further shows that the memory evolves consistently across layers: its similarity with current-layer features and the average similarity across layers both increase, indicating effective information accumulation.

These results suggest that Mema enables structured and progressive cross-layer interaction, rather than uniform feature smoothing.

\subsection{Ablation Study}

\noindent\textbf{\emph{Effect of Progressive Cross-layer Interaction.}}
To investigate the role of progressive cross-layer interaction, we compare Mema with variants that restrict or sparsify the interaction pathway (Table~\ref{tab:interaction_effect}).

Variants include applying interaction only at the last layer and sparsifying it across layers. 
While these settings provide partial improvements over the baseline, they consistently underperform the full Mema model. 
Restricting interaction to the last layer yields limited gains, and sparse interaction degrades performance, indicating that continuous cross-layer propagation is essential.

In contrast, Mema achieves the best performance by maintaining a persistent memory and applying interaction at every layer. 
These results demonstrate that effective cross-layer interaction must be both progressive and continuous to support hierarchical feature integration and alignment.

\begin{table}[t]
\centering
\setlength{\tabcolsep}{3pt}  
\small
\caption{
Effect of training data scale for Mema. We vary the number of training samples while keeping all other settings fixed. 
The \colorbox{oursblue}{blue row} indicates our default setting.
}
\label{tab:data_scale}
\begin{tabular}{lccccc}
\toprule
\textbf{Data Size} & \textbf{MME$^{P}$} & \textbf{MME$^{C}$} & \textbf{MMB} & \textbf{POPE} & \textbf{TextVQA} \\
\midrule
LLaVA-v1.5-7B & 1510.7 & 355.7 & 64.3 & 85.9 & 58.2 \\
\midrule
5K  & 1506.5 & 349.6 & 64.6 & 86.8 & 58.0 \\
10K & 1497.2 & 355.7 & 64.6 & 86.7 & 58.1 \\
\rowcolor{oursblue}
20K & \textbf{1520.6} & \textbf{372.2} & \textbf{64.7} & 86.7 & 58.2 \\
30K & 1487.0 & 364.6 & 64.3 & 86.8 & \textbf{58.3} \\
40K & 1506.1 & 365.0 & 64.3 & \textbf{87.0} & 58.1 \\
\bottomrule
\end{tabular}
\end{table}

\noindent\textbf{\emph{Ablation Study of Core Components.}}
Table~\ref{tab:ablation} presents an ablation study of Mema on LLaVA-v1.5-7B, including HME, textual conditioning, and MVA.
We consider three variants. 
\emph{w/o Text} removes textual conditioning from HME. 
\emph{w/o HME} removes the persistent memory mechanism, so each layer is modulated only based on its own pooled features without any cross-layer information.
\emph{w/o MVA} removes feedback refinement and directly adds the memory to the final visual representation. 
Removing textual conditioning results in slightly improved performance over the baseline on average, but still falls short of the full Mema model. This suggests that cross-layer interaction is effective on its own, while textual guidance further improves performance by steering memory updates toward task-relevant semantics.
Removing HME leads to performance close to the baseline, indicating that without a persistent memory, shallow-layer features cannot be effectively accumulated and propagated across layers, limiting the full utilization of fine-grained visual cues.
Removing MVA results in a significant performance drop, especially on TextVQA and MMB, indicating that memory must be actively injected into token representations to guide feature evolution across layers. Without this active feedback mechanism, shallow and deep features cannot effectively interact, hindering cross-layer alignment.

Overall, both HME and MVA are essential, with HME accumulating cross-layer information and MVA transforming it into effective token-level interaction.

\begin{figure}
    \centering
    \includegraphics[width=\linewidth]{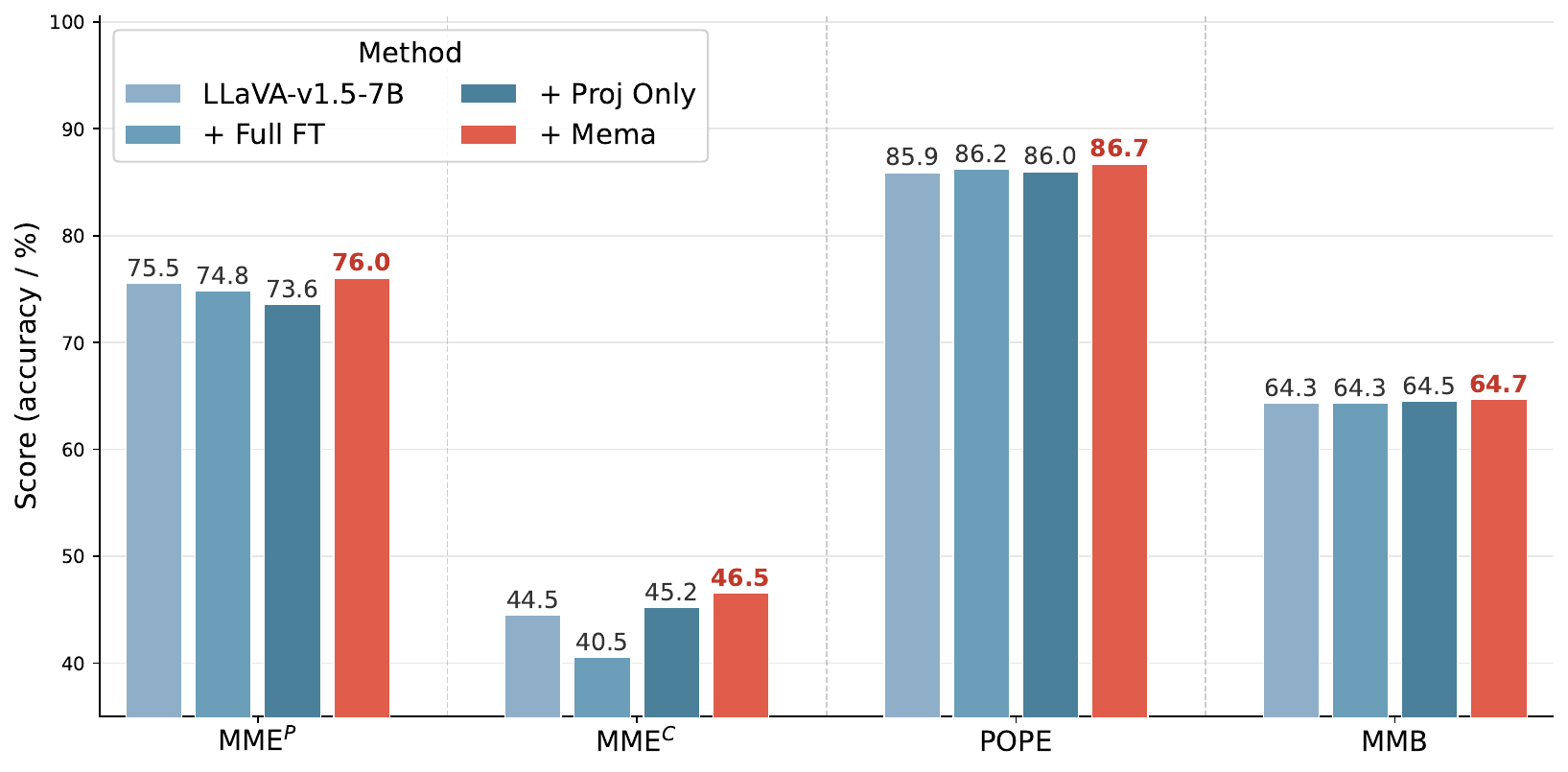}
    \caption{Effect of additional training data on the LLaVA-v1.5-7B. ``+ Full FT'' denotes fine-tuning both the LLM and the multimodal projector, while ``+ Proj Only'' denotes fine-tuning only the multimodal projector. }
    \label{fig:20k}
\end{figure}

\noindent\textbf{\emph{Sensitivity to Hyperparameters.}}
\label{sec:hyperparam}
We analyze the sensitivity of Mema to the reduction ratio $r$ and the alignment weight $\lambda$ in Table~\ref{tab:hyperparam}. 
We observe that a moderate reduction ratio ($r=4$) achieves the best overall performance, while both smaller ($r=1,2$) and larger ($r=8$) values lead to inferior results. 
This suggests that an appropriate bottleneck is crucial for stabilizing the memory update and preventing overly strong or weak feature modulation.

Similarly, $\lambda=0.1$ yields the most consistent performance, while both larger and smaller values degrade results, suggesting that an appropriate alignment strength is needed to preserve task-relevant semantics without over-constraining the representation.

\noindent\textbf{\emph{Data Efficiency Analysis.}}
To study the effect of training data scale, we vary the number of training samples used to train Mema, as shown in Table~\ref{tab:data_scale}. 
We observe that using a small number of samples (5K or 10K) leads to unstable or limited improvements, indicating insufficient supervision for learning effective cross-layer interaction. 
Increasing the data size to 20K yields the best overall performance across multiple benchmarks, suggesting that this scale is sufficient for learning robust interaction patterns. 
Further increasing the data size (30K and 40K) does not bring additional gains and even leads to slight performance degradation on some metrics. 
This suggests that Mema does not require large-scale retraining and can achieve strong performance with a relatively small subset of data.

\noindent\textbf{\emph{Effect of Additional Training Data.}}
To verify that the performance gains of Mema are not solely due to additional training data, we conduct controlled experiments using the same extra 20K samples on LLaVA-v1.5-7B. 

As shown in Fig.~\ref{fig:20k}, additional training yields limited and inconsistent improvements across benchmarks. 
Both baseline variants degrade performance on TextVQA and fail to improve MME. 
In contrast, Mema achieves consistent gains while maintaining stable performance across benchmarks.

These results suggest that the improvements brought by Mema are not simply due to additional data or extended training, but arise from the proposed cross-layer interaction mechanism, which improves representation evolution during encoding.

\subsection{Qualitative Analysis}

\begin{figure}
    \centering
    \includegraphics[width=\linewidth]{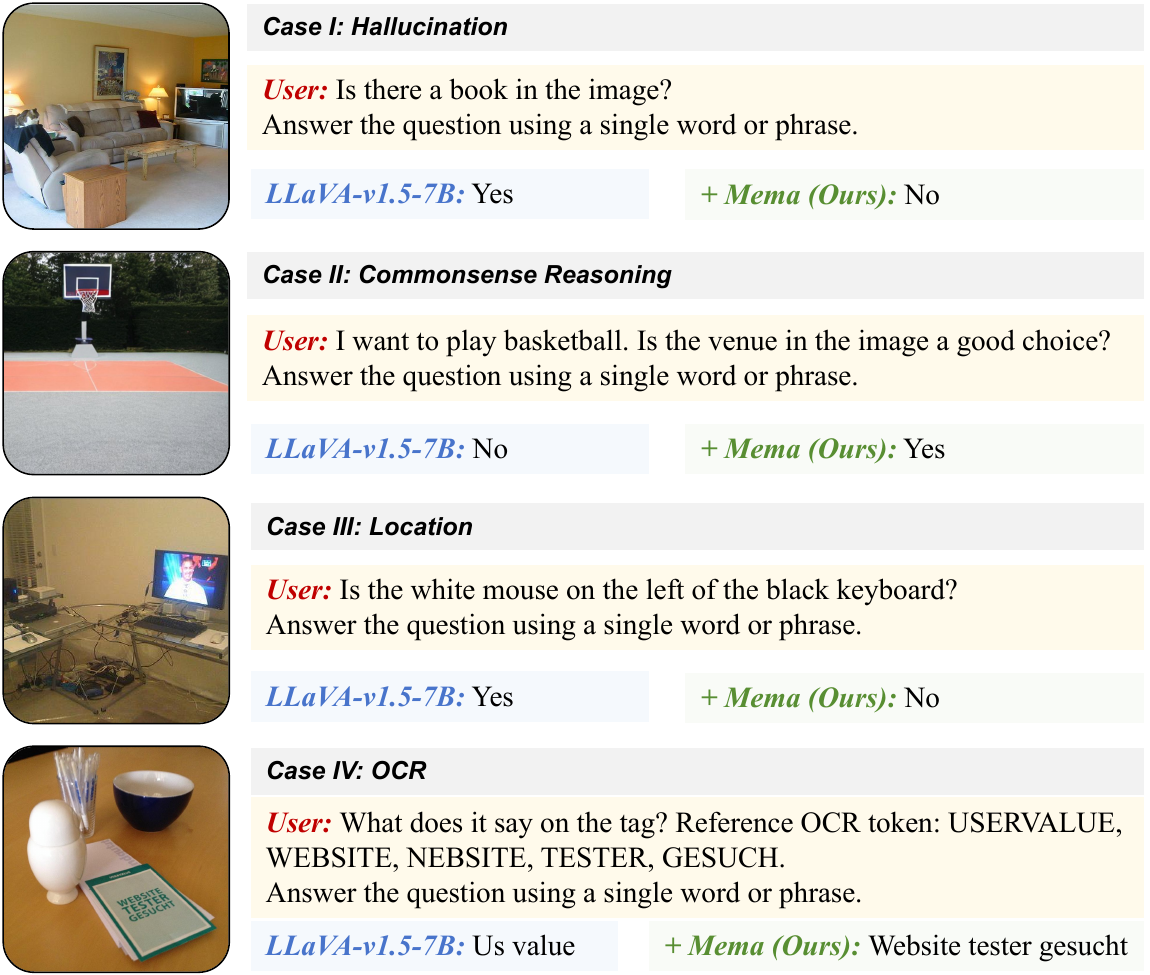}
    \caption{
    Qualitative comparison between LLaVA-v1.5-7B and Mema across diverse tasks (examples drawn from MME~\cite{mmb}, POPE~\cite{pope}, and TextVQA~\cite{textvqa}), including hallucination, commonsense reasoning, location, and OCR.
    }
    \label{fig:case-study}
\end{figure}

We provide qualitative comparisons between LLaVA-v1.5-7B and Mema in Fig.~\ref{fig:case-study}, covering diverse scenarios including hallucination, commonsense reasoning, spatial understanding, and OCR. 
Across these cases, Mema produces more accurate and reliable predictions, while the LLaVA-v1.5-7B exhibits errors such as hallucinated content, incorrect spatial relations, or incomplete text recognition. 

Specifically, Mema reduces hallucination by avoiding spurious visual interpretations, demonstrates improved commonsense reasoning by selecting contextually appropriate answers, and achieves more precise localization in spatial queries. In OCR scenarios, Mema is able to recover more complete and semantically coherent text compared to the baseline. 

These improvements suggest that Mema enhances the utilization of visual information across layers, enabling more robust visual grounding and better alignment between visual features and task semantics.

\section{Conclusion and Limitations}
In this paper, we address the under-utilization and misalignment of hierarchical visual features in MLLMs by introducing Mema, a memory-based mechanism for explicit and dynamic cross-layer interaction within the vision encoder. 
Our results show that such interaction allows shallow visual cues to be effectively preserved and integrated along with the depth iterations.

Extensive experiments demonstrate consistent improvements across diverse benchmarks. These findings suggest that modeling cross-layer interaction within the vision encoder provides a scalable and effective direction for improving visual representation quality in MLLMs. Additionally, Mema's low computational cost and parameter efficiency make it a highly practical solution for enhancing multimodal learning models.

\noindent\textbf{\emph{Limitations:}} 
Mema introduces additional computational overhead due to the memory and feedback mechanism compared to purely feed-forward architectures. 
Although this overhead is moderate, it may become more pronounced when scaling to deeper vision encoders or higher-resolution inputs. 
{
    \small
    \bibliographystyle{ieeenat_fullname}
    \bibliography{main}
}


\end{document}